\documentclass[DIV24,twocolumn,fontsize=9pt]{scrartcl}
\pdfoutput=1
\usepackage[T1]{fontenc}
\usepackage[utf8]{inputenc}
\usepackage[usenames, dvipsnames]{xcolor}
\usepackage[colorinlistoftodos]{todonotes}
\usepackage{float}
\usepackage[caption = false]{subfig}
\usepackage{amssymb}
\usepackage{graphicx}
\usepackage{pdflscape}
\usepackage{booktabs}
\usepackage[normalem]{ulem}
\usepackage{color,soul}
\usepackage[colorinlistoftodos]{todonotes}

\DeclareOldFontCommand{\bf}{\normalfont\bfseries}{\mathbf}

\usepackage{url}
\usepackage{palatino}
\usepackage{booktabs}

\usepackage[font={sf},labelfont={rm,bf,sf},small,]{caption}
\setcapindent{0cm}
\DeclareCaptionLabelSeparator{pp}{ | }

\usepackage[defernumbers=true,
  maxbibnames=1000,
  maxnames=30,
  backend=bibtex,
  firstinits=true,
  sorting=none,citetracker,
  url=false]{biblatex}
\renewcommand{\cite}{\supercite}
\DeclareFieldFormat{labelnumberwidth}{\mkbibbold{#1.}}

\newcommand\blfootnote[1]{%
  \begingroup
  \renewcommand\thefootnote{}\footnote{#1}%
  \addtocounter{footnote}{-1}%
  \endgroup
}

\usepackage{hyperref}

\usepackage{xcolor}
\hypersetup{
    colorlinks,
    linkcolor={red!50!black},
    citecolor={blue!50!black},
    urlcolor={blue!50!black}
}
\usepackage{enumitem}
\setitemize{noitemsep,topsep=0pt,parsep=0pt,partopsep=0pt}
\usepackage{balance}

\RequirePackage{fancyhdr}
\RedeclareSectionCommand[
  beforeskip=\baselineskip,
  afterskip=0.25\baselineskip]{section}
\usepackage{lettrine}

\begin{document}

\thispagestyle{fancy}
\pagestyle{fancy}

%\mainmatter 

\title{Evolving embodied intelligence from materials to machines}
%\titlerunning{Materials to Machines}

\author{David Howard
\and Agoston E. Eiben
\and Danielle Frances Kennedy 
\and Jean-Baptiste Mouret
\and Philip Valencia
\and Dave Winkler}
%
%\authorrunning{David Howard et al.}
% \institute{
% $^1$ The Commonwealth Scientific and Industrial Research Organisation (CSIRO), Australia\\
% $^2$ Vrije Universiteit Amsterdam, Netherlands\\
% $^3$ Inria, CNRS, Universit\'e de Lorraine, France\\
% $^4$ Latrobe Institute for Molecular Science, La Trobe University, Australia\\
% $^5$ Monash Institute of Pharmaceutical Sciences, Monash University Australia}

%\maketitle

\twocolumn[{
    \textbf{\textsf{This manuscript is the pre-submission manuscript provided by the authors.\\ For the final, post-review version, please see:\\ \url{https://www.nature.com/articles/s42256-018-0009-9}}}

    \bigskip

\hrule width \hsize \kern 0.5mm 
\hrule width \hsize \kern 0.5mm 
\hrule width \hsize height 0.5mm 

\begin{@twocolumnfalse}

    \begin{flushleft}
        \fontsize{23}{8}\selectfont
      \bfseries \sffamily
      Evolving embodied intelligence from materials to machines

      \bigskip

      \fontsize{12}{8}\selectfont

      David Howard\textsuperscript{1*}, 
      Agoston E. Eiben\textsuperscript{2}, 
      Danielle Frances Kennedy\textsuperscript{1}, 
      Jean-Baptiste Mouret\textsuperscript{3}, 
      Philip Valencia\textsuperscript{1}, 
      and Dave Winkler\textsuperscript{1,4,5}
    \end{flushleft}
\smallskip

\bfseries \sffamily Natural lifeforms specialise to their environmental niches across many levels; from low-level features such as DNA and proteins, through to higher-level artefacts including eyes, limbs, and overarching body plans.  We propose Multi-Level Evolution (MLE), a bottom-up automatic process that designs robots across multiple levels and niches them to tasks and environmental conditions.  MLE concurrently explores constituent molecular and material 'building blocks', as well as their possible assemblies into specialised morphological and sensorimotor configurations.  MLE provides a route to fully harness a recent explosion in available candidate materials and ongoing advances in rapid manufacturing processes.  We outline a feasible MLE architecture that realises this vision, highlight the main roadblocks and how they may be overcome, and show robotic applications to which MLE is particularly suited.  By forming a research agenda to stimulate discussion between researchers in related fields, we hope to inspire the pursuit of multi-level robotic design all the way from material to machine. 

\smallskip

\emph{Perspective article} --- Howard, David, et al. ``Evolving embodied intelligence from materials to machines.'' Nature Machine Intelligence 1.1 (2019): 12.

      \normalfont
  \end{@twocolumnfalse}

\vspace{1em}
  }]

\blfootnote{\sffamily \textsubscript{1} The Commonwealth Scientific and Industrial Research Organisation (CSIRO), Australia}
\blfootnote{\sffamily \textsubscript{2} Vrije Universiteit Amsterdam, Netherlands}
\blfootnote{\sffamily \textsubscript{3} Inria, CNRS, Universit\'e de Lorraine, France}
\blfootnote{\sffamily \textsuperscript{4} Latrobe Institute for Molecular Science, La Trobe University, Australia}
\blfootnote{\sffamily \textsuperscript{5} Monash Institute of Pharmaceutical Sciences, Monash University Australia}
\blfootnote{\sffamily *e-mail: \href{mailto:david.howard@csiro.au}{david.howard@csiro.au}}

%We propose Multi-Level Evolution (MLE), a bottom-up automatic process that designs robots across multiple levels and niches them to tasks and environmental conditions

\lettrine[lines=3]{\textsf{R}}{obots} are on the rise, and seen with increasing ubiquity in what are known as `structured' environments.  Pick and place machines are a good example; their interactions with the environment are predictable and so easily controllable.  As a counterpoint, robots consistently struggle in complex, unpredictable 'unstructured' environments \cite{carlson2005ugvs,atkeson2018}. Cataloguing biodiversity in remote areas, searching destroyed buildings for survivors following an earthquake, and exploring  labyrinthine cave systems are good examples.

%At first glance, it would appear that successfully operating in these environments would require highly complex, versatile robots that are ready for all situations.  Nature highlights an alternative path, as evolution specialises specialist creatures to thrive in incredibly challenging environmental niches \cite{rothschild2001life}.  
\section*{The challenge of embodied intelligence}

Natural life thrives in unstructured environments through a specific brand of intelligence known as \emph{embodied cognition} \cite{barrett2011beyond}.  Intelligent behaviour emerges from tight coupling between an agent's body, brain, and environment, not solely from the brain.  In the taxonomy of philosophy, it opposes the `I think therefore I am' of Descartes, and a plethora of research to date has shown that the form and function of an agents physical presence plays an important role in learning, development, and the generation of suitable in-environment behaviour \cite{pfeifer_how_2006}. 

Life's ability to produce useful embodiments comes from a free-form evolutionary process where variance occurs across multiple \emph{levels} \cite{carroll2013dna}. Generally speaking, mutations in low-level DNA lead to changes in protein expression, facilitating an emergent process defining the structure and composition of higher-level features including eyes, hands, and limbs, and their placement in body plans.  Making robot design similarly free-form and level-based might herald a new wave of capable embodiments to finally tackle challenging unstructured environments. 

%DESIGNING CAPABLE EMBODIED ROBOTS!!! 
Embodied cognition and its artificial analogue, embodied Artificial Intelligence (AI) \cite{brooks1991intelligence,pfeifer_how_2006}, have long known that complex environments can only be tackled by sufficiently capable combinations of body and brain.  In robots, body design has lagged behind due to inherent manufacturing complexities (through cell division, nature gets this ability `for free').  As such, including a variety of materials into robot design has been a long-standing `holy grail' for embodied AI \cite{pfeifer_how_2006} and robotics in general.  We present a straightforward, reasonably scaleable path towards incorporating material search and selection into robot design: a more free-form and specialising process than any currently available method. 

We call this algorithmic framework Multi-Level Evolution, or MLE.  Here we consider a three-level architecture, although as many levels as necessary may be instantiated depending on problem demands.  Three levels is a natural split, based on longstanding delineations across the large, active and well-established fields of materials science, robotics, and component design.  

At the lowest level, materials are discovered. Components are created by selecting one or more materials into a geometry.  Finally, robots are created by integrating components into template `body plans', and evaluating them on a task in an environment.   New candidate materials and components are discovered during the process, continually increasing the range of possible robot designs.

%MLE is inspired by three key facets of natural evolution: (i) variance occurs across multiple interrelated levels, providing freedom to find useful solutions and the opportunity to specialise (ii) changes in a lower level affect the emergence of features at higher levels, and (iii) specialisation is observed across a range of environmental niches. 

%COULD REMOVE?
A key takeaway is that MLE directly contrasts conventional engineering approaches which, because of time and cost constraints, often search for versatile generalist robots that do a bit of everything at a reasonable level of performance.  MLE is a universal designer (across a wider design space, and for any task-environment niche) that generates specialists by harnessing {\em diversity} across all levels, and the {\em emergence} of useful artefacts and artefact combinations.

In this Perspective, we outline how new types of artificial evolution, which are ready to exploit the same wave of ubiquitous computing resources that powered the rise of deep learning, and have been shown to achieve a corresponding leap in performance \cite{salimans2017evolution}, can harness the recent explosion of available materials and manufacturing techniques (see boxouts) to create powerfully-embodied robots. We review the main roadblocks to the realisation of this vision.  We define the key features of MLE architectures, and, using examples of cutting-edge evolutionary algorithms, propose a simple implementation. We highlight use cases to which MLE is particularly suited.  Finally, we sketch out a path towards increasingly capable MLE implementations, and discuss the implications of MLE for the field of robotics.

\noindent\fbox{\begin{minipage}{0.95\columnwidth}
\small
\section*{Enabling technologies: materials}
\sffamily

According to the laws of chemistry, the number of materials available for search is $\approx10^{100}$ (see ref\cite{le_discovery_2016}) (for comparison, there are `only' $10^{82}$ atoms in the universe.  This provides an almost infinite toolbox for designing bespoke, functional materials for robots, including new sensing, actuation, and power materials\cite{yang_new_2017}, that we are increasingly able to design, characterise, and synthesise.  Accelerating the development cycle of bespoke materials is key as the space of possibilities is so vast (using e.g., robotic materials synthesis\cite{soldatova_ontology_2006}, or combinatorial materials libraries\cite{maier_combinatorial_2007}).

Semi- or fully-autonomous `closed loop' systems use robots to efficiently perform experiments with reduced reliance on humans, and naturally couples with techniques that automatically plan optimal sets of experiments \cite{sans_towards_2016,king_automating_2015} to reach desired material properties\cite{granda_controlling_2018}.  Robots can perform multiple simultaneous experiments, vastly reducing time and human effort whilst increasing providence of experimental data for computational, AI and machine learning methods, which are now mature enough to reliably predict properties of new materials and allow a vast set of previously-physical experiments to be conducted (cheaper and faster) virtually \cite{le_discovery_2016}.  High-throughput computational techniques exploit massive data-sets and advanced modelling to the same effect\cite{curtarolo2013high,pyzer-knapp_what_2015}.

Combined, these advances provide unprecedented opportunities to design and manufacture `smarter', more specialised robotic materials that integrate sensing, actuation, and other properties to create powerful embodiment options \cite{menguc_will_2017}.

\normalsize
\end{minipage}}

\noindent\fbox{\begin{minipage}{0.95\columnwidth}
\small\sffamily

\section*{Enabling technologies: advanced manufacturing}
MLE is poised to exploit additive \cite{calignano_overview_2017} and subtractive manufacturing of free-form structures and complex geometries\cite{Li_AddvSub_2018}, printing intricate multi-part components from multiple materials in-situ.

Behavioural diversity can be embedded during the manufacturing process through functional gradation \cite{eujin_study_2017,martinez:hal-01697103} to vary material properties (e.g., stiffness, elasticity).  Voxel blending gives fine-resolution, gradual tuning of build properties through continuous mixing of multiple materials.  Production devices tuned for ever-expanding range of feedstock increases the diversity observed in recent composite and multi-material robots, e.g. \cite{chen2017integrated}. 

MLE is iterative, so streamlining construction and reducing reliance on humans is a priority.  Sensors, actuators, and power systems are readily printable in various configurations \cite{HAGHIASHTIANI20181}, and continue to close the performance gap with their traditional counterparts whilst being increasingly integrated into multi-function materials during construction \cite{menguc_will_2017}.  MLE is poised to benefit from (semi-)autonomous robot construction, including prototype generate-and-test systems \cite{rosendo2017trade,brodbeck_morphological_2015}, culminating with whole robots constructed without human intervention \cite{wehner_integrated_2016}.

%Relevant emerging technologies include 4D printing of shape-changing materials with powerful adaptive embodiment options \cite{Raviv_selfdeforming_2014,xin_intelligent_2017}, and programmed self-folding which offloads assembly effort onto the robot \cite{Yim_3DFOLDING_2018}.
\normalsize
\end{minipage}}

\section*{Inspiration and characteristics of Multi-Level architectures}

MLE is a natural extension of Evolutionary Robotics (ER); a field that harnesses iterative, population-based algorithms to generate robot bodies, brains, or both \cite{silva_open_2016}.  A typical ER experiment defines a representation; how the genotype (string of numbers) maps to a phenotype (physical robot).  To capture sufficient complexity, these representations are typically indirect - simple genotypes define more complex robot phenotypes, potentially incorporating naturally-observed features including gene reuse (e.g., to encode two identical eyes), radial and bilaterial symmetry (seen across nature in body plans), and scaling factors (across the five fingers of a hand)\cite{hornby2011computer}. It randomly initialises a population using the representation, and tests their task-environment performance against a user defined `fitness function'.  Analogies of genetic mutation and recombination induce variance in the genotypes to create a new generation, with a preference to select high-fitness parents.  This process iterates until some acceptable level of performance is met.  

ER provides environmental adaptation \cite{auerbach_environmental_2014}, and explores a wider design space than other approaches, locating unconventional `short-cut' designs which may by otherwise missed \cite{lehman2018surprising}.  
%%%CUTTABLE
Owing to a dearth of versatile, affordable manufacturing processes, ER traditionally focused on controller generation for fixed morphologies \cite{nolfi2000evolutionary}. Signalled by the first 3D printed evolved robot in 2000 \cite{lipson2000automatic}, we now find ourselves in the era of the `Evolution  of Things' \cite{Eiben2012Embodied-Artifi,eiben_evolutionary_2015}, where complex physical artefacts are evolved and physically instantiated \cite{rieffel2017introduction}.

From the MLE perspective, `classic' ER is the top-level level that finds environment and task specific controllers and body plans - arrangements of structure, sensing, and actuation that together comprise a robot.  Materials are not typically considered as part of the robot's `genome' (although idealised materials properties appear sporadically in simulation \cite{cheney_unshackling_2013}).  We posit that the missing link to unlocking richer embodiments is to discover, model, and select real (and newly-discovered) materials, and make them available in a holistic design process.  MLE architectures are characterised by:

%MLE advocates for holistic design architectures that discover materials and select them into components (and select components into body plans...).  

\begin{enumerate}
\item Three vertically-stacked levels (robot, component, material).  Robots are arrangements of components, where a component is a combination of a geometry and one or more materials that occupy sections of the geometry.  
\item At least one search process per level, which is responsible for finding new artefacts within a given level.  In the component level, we could run search processes for actuators, sensors, and structural elements.
\item Hybridisation, a novel concept that enforces that real and virtual genomes are identical for either physical or virtual instantiation.  This means we can easily `cross-breed' between simulated and real artefacts.
\end{enumerate}

We suggest evolutionary algorithms as bias-free and domain-agnostic\cite{eiben_introduction_2003} \emph{default} algorithms, with a track record of success in discovering molecules and materials \cite{le_discovery_2016}, components and structures \cite{stanley2007compositional}, and robots \cite{10.3389/frobt.2015.00004}, whilst being relatively efficient across all of these levels \cite{rabitz_control_2012}. As each level is independent, we can use domain specific algorithms/representations as requires.  For example, the materials level may benefit from capturing the underlying phenomena relating a materials structure to its behaviour \cite{hansch1995exploring}.  

%Similarly, MLE delivers three key features: (i) the ability to handle far more design variables than a human designer (including materials selection), (ii) a generalist design approach, applicable to any task-environment combination, and (iii) a design with scalability, collaboration, and re-use in mind. 

With the `grand vision' sketched out, let us now consider how emerging technologies can build simple prototype MLE architectures in the near future.

\begin{figure*}[ht]
\centering
\subfloat{\includegraphics[width=\textwidth]{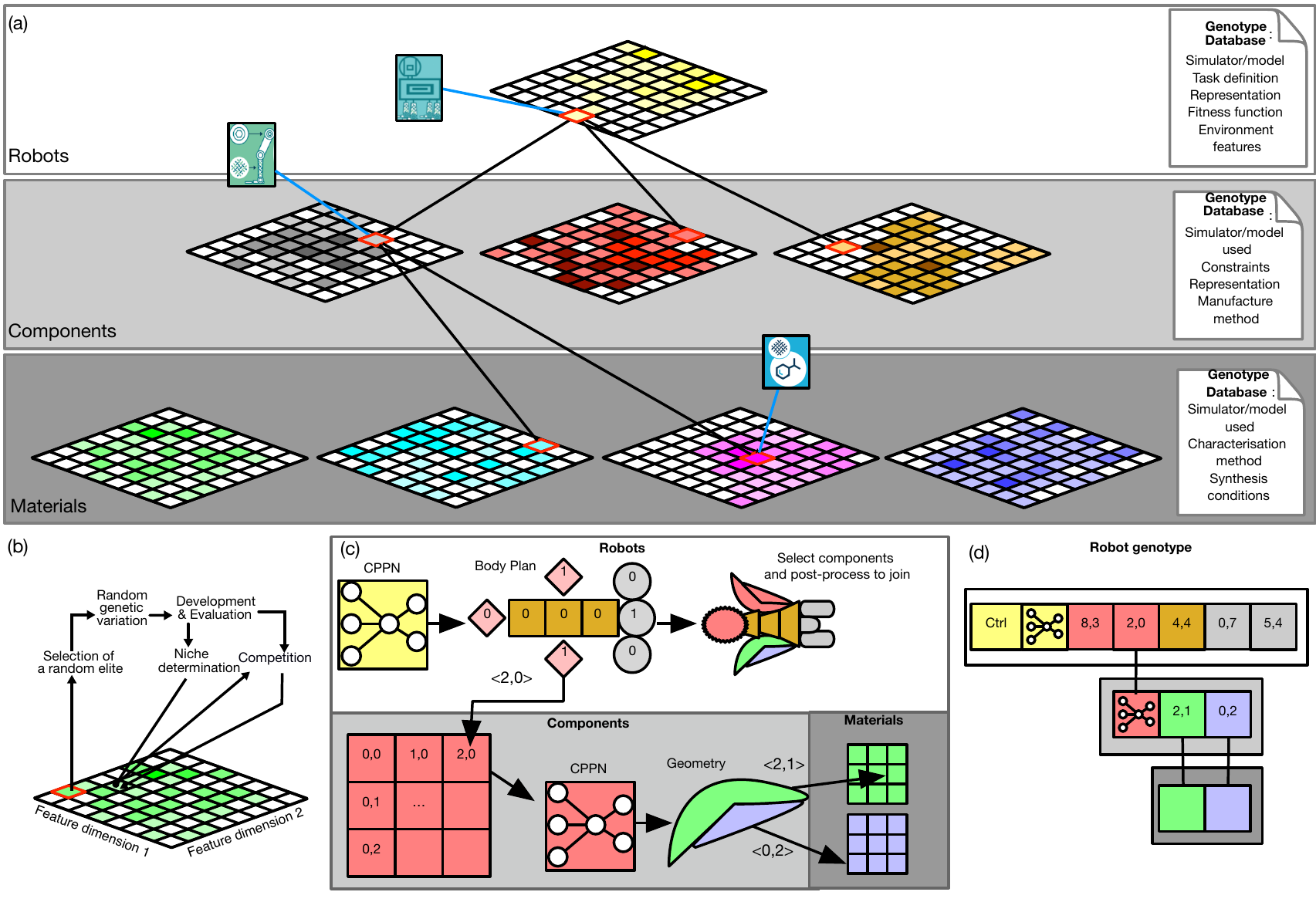}}
\caption{\textsf{\textbf{(a) A sample MLE architecture }incorporating a single robot search process, 3 component search processes (e.g., sensors, actuators, body segments), and 4 materials search processes (polymers, etc.).  For clarity only two dimensions of each process are shown, discretised into `bins'.  Colour indicates the highest fitness solution found per bin; darker represents a fitter solution, white squares indicate no current solution.  Over time, more bins are filled, and bin fitness is improved, which improves the quality and diversity of options available to upper levels. An asterisk (*) denotes an individual created physically, other individuals are virtual.  \textbf{(b) Creating a diversity of high-quality solutions.} At every iteration, the Illumination algorithm (1) randomly selects an occupied bin, (2) adds random mutations to the current best solution of the bin, (3) evaluates the quality (fitness) and the features of the newly generated solution, (4) compares the quality of the newly generated solution with the current best of the bin that corresponds to its features and keeps the best solution. These four steps are repeated until all the bins are occupied with satisfying solutions. \textbf{(c) Creating a robot.} At the top level, a Compositional Pattern Producing Network (CPPN)  defines the body plan.  Once generated, an appropriate number of pointers into the components layer are set based on the number of component slots generated, and the corresponding components fused via post-processing to create the final robot.  Pointers address a specific member of the library, in this case two integers for a two-dimensional library.   In this case components are segregated into actuation (red diamonds), body structure (gold squares) and sensing (grey circles).  A further CPPN per-component defines component geometry, and subsequently the number of material pointers required for that component.  Coloured grids (red, green, purple) abstractly represent the larger libraries seen in (a,b).\textbf{(d) A hierarchical genotype} that uses `pointers' to fully, efficiently define the robot. The robot controller ('ctrl') is defined at the top level. Background colours in (a) (c) and (d) signify different levels.}}
\label{fig:levels}
\end{figure*}

\subsection*{A conceptual MLE architecture}

Our conceptual prototype uses evolutionary Illumination algorithms \cite{mouret2015illuminating} (also called Quality-Diversity \cite{pugh2016quality}) to produce diverse libraries of high-performing potential solutions across three levels (Fig. \ref{fig:levels}(a)). Libraries are n-dimensional grids of possible combinations of some physical properties, discretised into `bins' \cite{vassiliades2018using}.  For example, all actuators transmit force and consume energy, all materials possess weight, rigidity, elasticity, and compliance that can be exploited to generate robots that are adapted to specific environments.  By measuring these properties we can assign to the appropriate bin.  

For clarity, Fig.\ref{fig:levels} visualises two properties per process, resulting in 2D grids.  Practically, there will be many more properties (Table~\ref{table1}).  It is critical that MLE generates a diverse set of solutions, rather than a single optimal solution as in traditional optimization and classic ER.  Libraries allow each level to be explored independently and provides diversity to upper levels.  Each level may be subdivided; e.g., components can be subdivided into sensors, actuators, and body structure.  Each search process can also have its own solution representation, feature dimensions, and search operators.

\begin{table*}
    \renewcommand{\arraystretch}{1.6}
    \caption{\label{table1}
    \textsf{Sample search processes, tunable variables, and desired properties that may be found in each level (non-exhaustive).}}
\begin{small}
    \sffamily
\begin{center}
\begin{tabular}{p{1.5cm}p{3cm}p{6.25cm}p{5.25cm}} 
\toprule
\textbf{Level} & \textbf{Search processes}  & \textbf{Tunable variables}   & \textbf{Desired properties} \\ 
\midrule
Material 
& Polymer, metal, composite, nanomaterial, hydrogel, thin-film, shape-memory, dielectrics. 
& Level of experiment, structures and physiochemical properties (of reactants, reagents, substrates, solvents, catalysts, additives, and products), pH, particle size, shape and density, surface roughness and chemistry, pressure, temperature, and time, speed of mixing or coating, rate and order of addition; and post-processing conditions such as drying, conditioning or activation. 
& Flex, deformation, weight, response strength, energy requirements, hysteresis, signal/noise (sensing), repeatability, power density, peak voltage/current, useful temperature range, cost \\
\midrule
Component 
& Sensor, actuator, body structure, energy, multi-function			
& Geometric properties, material composition, gradation
& Load bearing,  strain density, impact resistance, weight, deflection, range of motion, force transmission, torque, compliance, flexibility, sensitivity (to light, chemicals, angular movement, etc...), power requirements, footprint, power provision \\
\midrule
Robot	
& Targeted locomotion, gripping, carrying/transporting, food seeking, recharging	
& Selected components, arrangement of components in body plan, controllers, post-processing algorithm
& Embodied behaviour (from control/robot/environment interactions) \\
\bottomrule
\end{tabular}
\end{center}
\end{small}
\end{table*}

%{\bf Representation types, direct or indirect.  CPPNS L systems SMILES, QSAR, etc. Pointers!!!}

To begin an experiment, we bootstrap the lowest level with known materials, either from the literature, or from previous MLE experiments.  Each material is placed in a bin based on its physical properties.  The components level then defines geometries and selects an appropriate number of materials into those geometries to create body segments, sensors, and actuators, filling some component bins.  At the highest level, we search for controllers and body plans; templates that define arrangements of components.  Here, geometries and body plan layouts are defined using Compositional Pattern Producing Networks (CPPNs), specialised neural networks evolved to output geometric patterns displaying modularity, regularity and symmetry, see e.g. \cite{StanleyNMI19}.

As well as belonging to a bin, each material, component, and robot has an associated fitness.  For materials and components, we suggest fitness based on the universally beneficial property of cost, therefore the cheapest example that fulfils certain physical property requirements will be passed to the next level to reduce the manufacturing burden.  More specific fitness measures, e.g., efficiency for an actuator, or signal-to-noise for a sensor, will be mediated by the environment, so we lose transferrability for potential gains in performance.  Robot fitness is based on its behaviour; how well it completes the task.

The grids at each level progressively fill out as new material, component, and robot designs are discovered (Fig. \ref{fig:levels}(b)). Illumination search specifically encourages diversity \cite{mouret2015illuminating} through pressure to discover new combinations of physical properties, providing larger libraries and thus more opportunities to exploit materials and components in interesting ways, facilitating emergent behaviour.  As lower levels focus on creating a diversity of options, significant opportunities arise for the spontaneous emergence of component-material combinations that facilitate useful behaviour, which will likely result in a good fitness score for the robot, with no constraints on exactly how that behaviour emerges.  These behaviours are a holistic combination of the search efforts at every level, and the in-environment performance of the resulting robot.  Counter-intuitively, Illumination search is known to discover more `optimal' outcomes than pure optimisation approaches \cite{mouret2015illuminating}; hence we expect high-performance artefacts.  Cascading improvements may percolate through the levels; when a new material is found it could improve the fitness of the solution in a populated bin (replacing the previous best), or it may expand the number of filled bins in its level, and potentially the number of reachable bins at any level above it.  

To instantiate a robot (\ref{fig:levels}(c)), we query the corresponding CPPN and seamlessly integrate the relevant components into the resulting body plan using post-processing.  Accompanying the CPPN are a number of `pointers' to bins in the components level, which selects specific components into the body plan.  Each pointer addresses a specific bin in the level below.  Similarly, a component consists of a geometry-defining CPPN and pointers to materials.  Either the CPPN or the indices of the pointers may be altered during evolutionary search, which changes the shape or composition of the affected artefacts.  Materials may be represented and searched in a similar way.  Once instantiated, the robot is evaluated based on desired mission performance to ascertain its fitness.  

Unlike natural genotypes, which are defined at the DNA level, the genotype of a robot produced by MLE can be thought of as a hierarchy \ref{fig:levels}(d), where the robot body plan and controller are defined at the top level.  Following the pointers from robot to components allows us to fully define the components used, and following each component's pointers allow us to fully define the materials that comprise each component.   

The only necessary inter-level communication is the passing of candidate solutions upwards for use by the next level.  For efficiency, and for ease of integration into higher level simulators/models, only phenotypic properties (i.e., of the physical solution created) are passed between levels.  The representation of a solution, plus details of experimental procedures, models/simulators used, learning algorithms, and evaluation tests are stored in a database by the relevant layer as required so results are repeatable.

Not all physical properties will be relevant in all situations, and can be safely ignored.  Sparse feature selection methods \cite{figueiredo2003adaptive,tibshirani1996regression}, applied as automatic dimensionality filters, give more weight to the features most relevant for fitness and function within the niche and minimise combinatorial issues.  In our example, this may select the physical properties in which we encourage diversity.  Similarly, not all options will be required from lower levels.  Each level can also design its own library from lower-level libraries according to its own objectives\cite{cully2018hierarchical}.  Combined, these processes reduce the required computational effort, and the extent of physical characterisation required.  

\subsection*{Learning and behaviour}

The focus of this Perspective is in improving the bodies of embodied robots.  However, we need some way of generating useful behaviour from these bodies.  In some cases, this can result solely from the interactions of materials and components in the robot’s body\cite{cheney_unshackling_2013}, or through automatic response to stresses experienced between body and environment\cite{kriegman2018interoceptive}.  This \emph{morphological computing} offloads the computation of behaviour from a controller onto the robot’s body\cite{hauser_towards_2011}, reducing the required controller complexity.  We expect MLE to greatly benefit from morphological computing, owing to the vast range of physical behavioural responses it can instantiate.

For more complex tasks, learning will be required to overtly direct the body-environment interactions our robots produce \cite{eiben2013triangle}, creating a controller --- the `ctrl' seen in the robot genome in Fig~\ref{fig:levels}(d).  Software provides a wealth of options including neural networks, central pattern generators, behaviour trees, and modular architectures \cite{silva_open_2016}, which can be optimised through reinforcement learning, evolutionary algorithms, and imitation.  Post-deployment \emph{online} learning offers the possibility to adapt controllers following hardware failures\cite{cully2015robots}, or to changing environmental conditions.  Ultimately the choice of controller and learning is a design decision, key requirements are that the body is controllable, its material and morphological composition properly exploited, and its behaviour suitable for our task.

\section*{Physical and virtual testing provides the best of both worlds}

Manufacturing and testing each new material, component, and robot in reality would be prohibitively expensive in terms of time and cost.   MLE's success hinges on effective use of simulation and modelling, and blurring the lines between real and virtual. 

MLE introduces the novel concept of \emph{Hybridisation}, such that the representation describing a material (or component, or robot) is identical, regardless of whether it is real or virtual. Bins in each level may be filled through physical experimentation, or the results of simulation or predictive model.  Simulated evolution runs concurrently with physical experimentation, and cross-breeding allows physical or virtual materials (or components, or robots) to parent a child that may exist in the real world, in the virtual world or in both. The advantages of hybridisation are significant; physical evolution is accelerated by the virtual component that can run faster to find good robot features with less time and fewer resources, whereas simulated evolution benefits from the influx of `genes' that are tested favourably in the real world. 

Physical experimentation provides necessary `ground truth' data, the burden of which may be reduced through smart algorithmic design \cite{bhattacharya_evolutionary_2013}.  Evaluating robot performance in reality is particularly difficult (Repeatability and physical damage are key problems), but increasingly feasible thanks to custom-designed test arenas \cite{howard_platform_2017,7989128}, and proof-of-concept `generate and test' facilities  \cite{brodbeck_morphological_2015}. 

We must be able to simulate the performance of constituent materials and components in the top-level robot.  Where possible, conducting all evaluations in the same simulator guarantees interoperability; nearly all simulators allow various materials properties to be defined and directly specified from lower levels.  Multiscale modelling can enhance veracity.  Some properties, for example hyperelasticity, can be tricky to model and may require specialist tools. In this case,  co-simulation can be used to link specialist simulators together, allowing materials to be simulated, and their results shared with a dedicated component simulator to determine overall performance.  Such approaches integrate with techniques that automatically validate the material models for use in simulators\cite{back2011evolutionary}. Directly representing complex micro-level behaviours in higher-level models/simulators is difficult, but increasingly feasible as it receives ongoing research attention in multiple fields in materials science.  %Multi-physics modelling allows our robots to be accurately modelled, e.g., thermodynamic, electromagnetic, and integrating these effects will give an accurate holistic view of their behaviour.

A key issue is the \emph{reality gap}, where necessary abstractions lead to degraded performance when simulator-designed artefacts are transferred to reality.  MLE heavily exploits techniques to reduce the gap.  Gathering data on real designs and using a learning algorithm (e.g., a neural network \cite{husken2005structure} or Gaussian processes \cite{williams2006gaussian}) to create surrogate models of the performance \cite{jin2011surrogate} improves accuracy and speed, and has been successful at the material level \cite{winkler2017performance}, the design level \cite{gaier2018data}, and for robot controllers \cite{chatzilygeroudis2017iros}.  Physical testing can improve an existing simulator to more closely match reality, either tuning simulator parameters \cite{bongard2006resilient,zagal2007combining} and/or combining the predictions of the simulator with those of a data-driven model \cite{chatzilygeroudis2018icra}. In between these two ideas, it is possible to learn a `transferability function' that predicts the accuracy of the simulator for a given design \cite{koos2013transferability}.  %MLE is poised to benefit from current research directions in 'Sim-to-real', e.g.\cite{sadeghi2017sim2real}, and adapt domain randomisation\cite{james2017transferring} to further reduce the gap.

\subsection*{The Benefits of MLE}

MLE is primarily designed to harness materials to provide diverse, specialised robot designs.  Other main benefits include:
%MLE is designed to provide the following beneifts:

\begin{itemize}
    \item \textbf{Scalability:} Promoting scalability is necessary due to combinatorial issues \cite{donoho_high-dimensional_2000} brought about by embedding multiple search processes across three levels.   Distributing the `full genome' of a robot across multiple independent automatic design processes allows each level to be searched in parallel, using the specialist tools of each field where applicable to improve efficiency.  Hybridisation shifts the majority of the search effort into relatively cheap, parallelisable simulations and models.
    \item \textbf{Self-optimisation:} Although the early stages of MLE are likely to be slow, with few options available, we envisage the system as somewhat self-optimising; the longer it runs, the better our models become, and the more options are discovered in every layer.
    \item \textbf{Re-use:} Focusing on physical properties allows materials and components to transfer between MLE architectures.  Processes can be swapped in or out of a level with relative ease.  
    \item \textbf{Collaboration:} MLE architectures will likely be distributed across multiple institutions depending on the availability of hardware and specialists, leading to an inherently collaborative effort integrating multiple research groups and the architecture itself, that promotes standardised, readily available experimental information and cross-fertilisation of ideas\cite{wagy2015combining}.  We may look to the Materials Genome Initiative for inspiration on standardising MLE, encouraging collaboration, and reducing barriers to entry \cite{jain2013commentary}.
\end{itemize}

\section*{Opportunities for MLE architectures}
As a new paradigm for designing robots, MLE will naturally gravitate towards certain applications.  Consider the rapidly-advancing field of soft robotics \cite{tolley_resilient_2014,rus_design_2015};  compliant, deformable robots that survive crushing, burning, and other hazards which are characteristic of the unstructured environments we want to put robots into.  Integration of sensing, actuation, and deformation are fundamental to soft robotics; MLE most simply permits this using a single library of multi-function components, rather than dedicated sensing, actuation, etc. 

Soft robotics currently lacks a codified design methodology \cite{lipson_challenges_2014}, as deformable soft materials are not amenable to conventional approaches \cite{bongard_evolving_2016}.   Designers often settle on a (frequently bio-inspired) preconceived design, for example an octopus or a jellyfish \cite{yeom_biomimetic_2009}, which instantly places heavy limitations on the designs considered.  Rather than design a fish, MLE lets us ask a different question - what `creatures' might evolution devise if its building blocks were not protein, muscle, and bone; but rather polymers and composites?  MLE is a perfect fit for the role of soft robot designer, harnessing diversity to comprehensively explore soft robot design spaces.  

Soft robots have particularly interesting and powerful embodiments, which emerge through interacting arrangements of morphological and material properties \cite{manti_stiffening_2016,bauer201425th}. MLE provides a continuous stream of new materials and increasingly capable componentry \cite{han_overview_2017,miriyev_soft_2017}, offering a pathway towards \emph{designing for embodiment}: discovering specialized soft materials, fully leveraging those materials through the emergent generation of components and bodies, and creating controllers to strongly couple the resulting embodiments with the environment.  

%{\bf Sof trobot comment on components... Components are not specifically part of a body plan, but a useful abstraction.  Go back to sensibly picking the design parameters...   modularity depush! NOT THE SAME AS A MODULAR ROBOT, TYPICALLY SIMILAR BUILDING ZBLOCKS< THIS LETS US EFECTIVELY PUT SENSORS AND ACTUATORS ANYWHERE ON THE ROBOT WITH PATTERNING...}

For our second design opportunity, let’s cast our minds forward 20-30 years. Imagine that we want to perform basic environment monitoring with robots: to traverse terrain in a zone, gather some data, and fully degrade after a while. This might sound simple at first, but critically depends on the environment. The Sahara is very hot, dry, and sunny (during the day), but Antarctica is cold and icy.  Creepers and other low-lying foliage in the Amazon present a markedly different challenge to rolling desert sand dunes.  

%No generalist robot would be fit for all of these environments; too much standardisation leads to degraded performance, especially when design objectives start to contradict each other.  In an attempt to be 'ready for anything', each of these robots is likely to be more expensive to manufacture, and more difficult to replace, than a basic, specialised robot.

\begin{figure}[ht]
\centering
\subfloat{\includegraphics[width = \linewidth]{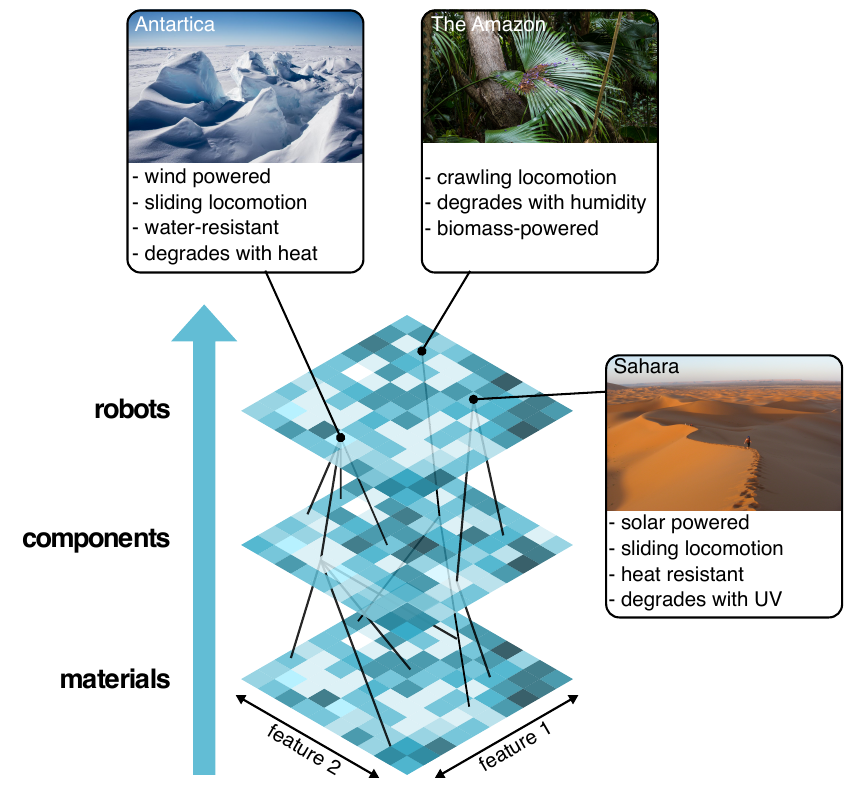}}
\caption{\sffamily \textbf{Showing how MLE can provide a diversity of robots for a diversity of environmental niches.} In this alternate MLE architecture, each level consists of only one heterogeneous library of solutions.  Such architectures are likely to be possible in the further future, where boundaries between sensing, actuation, and structure are collapsed to promote emergence and integration, at the cost of increased computational costs and combinatorial effort.
}
\label{fig:JBM}
\end{figure}

Designing robots for each niche with classic engineering would require an army of engineers for each environment, and the engineering cost would sky-rocket. This is why most of engineering is about standardization and not specialization.  The alternative is MLE (Fig.\ref{fig:JBM}), which could automatically design suitable robots (unique combinations of materials, morphology, and behaviour) for each environment. They might resemble insects: relatively simple, small, highly integrated, highly specialized, and fit for function.  Note that the same MLE architecture, with shared materials and identical objectives, could adapt robots to account for seasonal differences within a biome, or could design for each of the following environments, and provide the following features:

\begin{itemize}
\item Antarctica: wind-powered, sliding locomotion, water-resistant, degrades with heat (in the summer)
\item The Amazon: crawling locomotion, degrades with humidity, bio-mass powered
\item Sahara: solar-powered, sliding locomotion, heat resistant, degrades with UV
\end{itemize}

%They would typically have a short lifetime; if the environment or the objective changes, they would degrade and would be replaced by new designs. But they would not necessarily be capable of quickly adapting to changes during their life (like humans do).  

\section*{Towards a new era of Embodied Intelligence}

The convergence in materials, manufacturing, and design paves the way towards radically new ways of producing robots.  The main thesis of this Perspective is that MLE architectures can integrate different technologies and levels under an evolutionary umbrella.  Considering that natural evolution succeeded in filling practically all environmental niches on Earth with highly adapted lifeforms, this approach holds great promise as a robotic design technique.  MLE is admittedly ambitious, and as such we have identified four key challenges to be overcome during an MLE research program.

\begin{itemize}
    \item Initial designs will be constrained to materials that are easily created, characterised, and modelled.  MLE materials search, together with high-throughput efforts globally, and advances in materials modelling will gradually alleviate this issue.
    \item Efficiency: In an attempt to create emergent embodiments by filling as many bins as possible, our conceptual prototype trades diversity for efficiency.  To counter, imagine an `overseer' program that greedily searches for good embodiments, allowing pointers to {\em any} material properties rather than experimentally confirmed or modelled properties, and subsequently skewing the search process to find materials with those properties if a promising embodiment is found.  This leads to another issue around balancing the tension between any top-down influence and a bottom-up robot design process.
    \item Allocating resources across levels, e.g., including enough physical experimentation to keep simulations approximate to reality, which may be quantified and managed using Gaussian Processes to identify areas of uncertainty in our models.   Bottlenecks are another issue; insufficient resource allocation to e.g., actuators may limit the range of final robot design. Using discretised libraries lets us estimate coverage, and allocate more resources to searches lacking coverage.
    \item Ideally, MLE would be fully autonomous.  However the human designer will play a significant role for the foreseeable future; setting up (designing suitable measures of robot fitness, suitably discretising libraries, identifying suitable materials...) and running experiments (characterising materials, assembling and evaluating robots).  Ongoing developments in automated characterisation, testing, and construction facilities will reduce this burden.  %Note that this is distinct from the ideal of collaborative robot-producing systems; here we mean humans having to expend effort to perform tasks to keep MLE running, rather than beneficially interacting with each other and the MLE process \cite{wagy2015combining}.  
\end{itemize}

As well as challenges, we wish to highlight a significant opportunity; \emph{representations} (recall that representations are mappings from genotype to phenotype).  The main historical event in ER was a move from direct (1-to-1 mappings) to indirect representations, a response to increased phenotypic complexity required in real-world artefacts.  Hierarchical representation has received scant consideration to date; we see a huge opportunity for intelligent level-spanning representations, to describe complex artefacts \emph{that are built from other artefacts} and their interactions.

Looking to the future, we envision a staged development of MLE systems:

\begin{itemize}
\item Stage 1. Within 5 years, prototype MLE systems will come online, spread across multiple research institutions. They will produce evolved robots in controlled lab settings, with heavy human intervention.  
\item Stage 2. In about a decade, MLE systems will be able to generate robots for a suitably constrained real-world mission. Increasingly integrated construction techniques will speed up evolution and reduce the amount of human intervention required.
\item Stage 3, after around 20 years, will see deployments in real-world environments. As models become more sophisticated, and computing power more available, monolithic search processes will begin to merge, heightening the interplay between material and morphology and encouraging emergence (e.g., Fig~\ref{fig:JBM}).
\end{itemize}

Somewhat counter-intuitively for an architecture based on segregated levels, MLE is about collapsing boundaries; between research institutions, between scientific disciplines, between reality and virtuality, and between robots and their constituent materials.  In doing so, we hope to create a holistic design process for a new type of robot, specialised all the way from material to machine.

\section*{Competing Interests}
The authors declare no competing financial and non-financial interest.

\section*{Acknowledgements}
DH, DFK, PV and DW would like to acknowledge Active Integrated Matter, one of CSIRO’s Future Science Platforms, for funding this research. JBM is funded by the European Union's Horizon 2020 research and innovation programme (grant agreement number 637972, project ``ResiBots'').

%\bibliographystyle{naturemag}  
%\bibliography{bib/refs.bib}
\balance
\begin{small}
\printbibliography
\end{small}
\end{document}